\documentclass[11pt,a4paper]{article}
\usepackage[hyperref]{acl2021}
\usepackage{times}
\usepackage{latexsym}

\usepackage{microtype}
\usepackage{array}

\aclfinalcopy 
\def\aclpaperid{2138} 

\usepackage{amsmath}
\usepackage{amssymb}

\usepackage{adjustbox}

\usepackage{url}
\usepackage{amssymb}
\usepackage{bbm}
\usepackage{multirow}
\usepackage{stfloats}
\usepackage{graphicx}
\usepackage{booktabs}
\usepackage{xspace}

\usepackage{bbding}
\usepackage{pifont}
\usepackage{wasysym}
\usepackage{amssymb}
\usepackage{nccmath}
\usepackage{enumitem}

\newcommand{\cmark}{\ding{51}}
\newcommand{\xmark}{\ding{56}}

\DeclareMathOperator*{\argmax}{argmax}

\newcommand{\methodname}{SEQA\xspace}

\title{A Semantic-based Method for Unsupervised Commonsense \\Question Answering}

\author{Yilin Niu$^1$\footnotemark[1] , Fei Huang$^1$\footnotemark[1] , Jiaming Liang$^1$, Wenkai Chen$^2$, Xiaoyan Zhu$^1$, Minlie Huang$^1$\footnotemark[2]\\
  $^1$ The CoAI group, DCST; $^1$ Institute for Artificial Intelligence;\\
  $^1$ State Key Lab of Intelligent Technology and Systems;\\
  $^1$ Beijing National Research Center for Information Science and Technology;\\ 
  $^1$ Tsinghua University, Beijing 100084, China. \\
  $^2$ School of Computer Science and Technology, \\
  Beijing University of Posts and Telecommunications \\
  {\small \tt niuyl14@tsinghua.org.cn}
  \quad {\small \tt \{f-huang,liangjm18\}18@mails.tsinghua.edu.cn}\\
  {\small \tt wkchen630@gmail.com}
  \quad {\small \tt \{zxy-dcs,aihuang\}@tsinghua.edu.cn}}

\date{}

\begin{document}
\maketitle

\renewcommand{\thefootnote}{\fnsymbol{footnote}}
\footnotetext[1]{Equal contribution}
\footnotetext[2]{Corresponding author: Minlie Huang.}
\renewcommand{\thefootnote}{\arabic{footnote}}

\begin{abstract}
  Unsupervised commonsense question answering is appealing since it does not rely on any labeled task data. Among existing work, a popular solution is to use pre-trained language models to score candidate choices directly conditioned on the question or context. However, such scores from language models can be easily affected by irrelevant factors, such as word frequencies, sentence structures, etc. 
  These distracting factors may not only mislead the model to choose a wrong answer but also make it oversensitive to lexical perturbations in candidate answers.
  
  In this paper, we present a novel SEmantic-based Question Answering method (\methodname) for unsupervised commonsense question answering. Instead of directly scoring each answer choice, our method first generates a set of plausible answers with generative models (e.g., GPT-2), and then uses these plausible answers to select the correct choice by considering the semantic similarity between each plausible answer and each choice. We devise a simple, yet sound formalism for this idea and 
 verify its effectiveness and robustness with extensive experiments.
  %
  We evaluate the proposed method 
  on four benchmark datasets, 
and our method achieves the best results in unsupervised settings.
  Moreover, when attacked by TextFooler~\citep{TextFooler} with synonym replacement, 
  \methodname demonstrates much less performance drops than baselines, thereby indicating stronger robustness.
\end{abstract}

\section{Introduction}

\begin{figure}[ht]
    \centering
    \includegraphics[width=\linewidth]{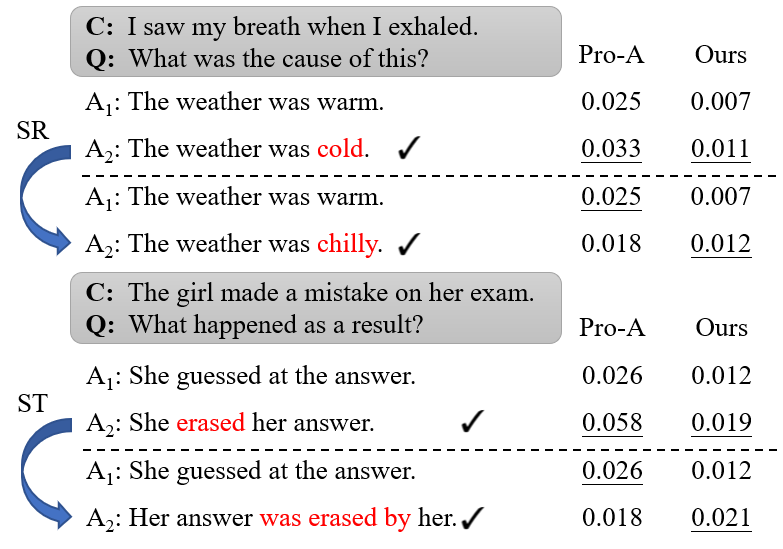}
    \caption{Two examples of commonsense question answering, where the baseline (Pro-A) is oversensitive to lexical perturbations (SR for synonym replacement and ST for sentence structure transformation). The scores from Pro-A and our method for each answer choice are shown in the right columns. The underlined score indicates the answer choice selected by a method.}
    \label{fig:sensitive-examples}
    \vspace{-1.5em}
\end{figure}

Pre-trained language models have been widely used for commonsense question answering. Finetuning pre-trained models on task-specific data produces many state-of-the-art results~\citep{PathGenerator,UnifiedQA,KagNet}. However, this requires amounts of labeled task data. Therefore, it is vital to study unsupervised commonsense question answering without relying on any labeled downstream task data.
In this paper, we investigate multiple-choice commonsense question answering tasks in an unsupervised setting: given a question and a set of answer choices, a model is required to predict the most reasonable answer choice for the question, but without access to any labeled task data. 

Many existing unsupervised methods tackle these tasks by scoring each answer choice using a language model, e.g., estimating the generative probability of the answer choice conditioned on the question~\citep{ASimpleMethod,selfTalk,dynamicReasoner,LMisAllYouNeedForCSReasoning}.
Table \ref{tab:existing-methods} lists several typical score functions.
However, these scores can be easily influenced by word frequencies, sentence structures, and other factors, which can mislead the models and make existing methods oversensitive to lexical perturbations~\citep{SensitivityofLanguageModels,LMisAllYouNeedForCSReasoning}. Figure~\ref{fig:sensitive-examples} shows two examples. The correct choices are paraphrased via synonym replacement or structure transformation. In these examples, the baseline (Pro-A) produces much lower scores for the paraphrased choices and chooses the wrong choices.

Since existing methods can be easily distracted by irrelevant factors such as lexical perturbations,
we argue that a commonsense question answering method should \textbf{focus on the answers' semantics and assign similar scores to synonymous choices}.
To this end, we introduce a novel SEmantic-based Question Answering model, \methodname, which aims to robustly select correct answers in multi-choice commonsense question answering in an unsupervised setting. 
Instead of directly scoring an answer choice, we calculate the probability of observing the choice's semantics.
A choice's semantic score can be obtained by summing the generative probabilities of sentences that have the same semantic meanings with the choice, where the sentences are called the choice's \textit{supporters}.
However, it is hard to obtain the \textit{supporters} which have exactly the same semantic meanings with the choice, so we reformulate the semantic score into a soft version as explained in Section \ref{sec:method-description}. Each \textit{supporter} is weighed by the semantic similarity to the answer choice, which can be computed with some off-the-shelf models, such as SentenceBERT \citep{sentenceBERT}.
Since the \textit{supporters} and their weights depend on the semantics rather than the surface form of the answer choice, by this means, the effects of the distracting factors can be largely suppressed.
Moreover, synonymous choices are likely to share the same set of \textit{supporters}, so their scores are expected to be stably close.
Our contributions in this paper are summarized as follows:
\begin{itemize}[leftmargin=1em]
    \setlength{\itemsep}{0ex}
    \setlength{\parskip}{2px}
    
    \item We propose a semantic-based question answering model (\methodname) for robust commonsense question answering in an unsupervised setting. Instead of directly scoring the answer choices, our method first generates some plausible answers and then uses them to select the correct choice by considering the semantic similarity between each plausible answer and each choice. 
    
    \item We conduct experiments on four commonsense question answering datasets, where \methodname achieves the best performance compared with strong baselines. When attacked by TextFooler~\citep{TextFooler} with synonym replacement, our method performs remarkably more robustly. 
    
    
\end{itemize}

\section{Related Work}\label{sec:related-work}

Previous work has explored pre-trained language models (LMs) for unsupervised commonsense question answering. In general, these approaches treat LMs as question answering modules.

Table~\ref{tab:existing-methods} shows three representative methods, which do not use external knowledge and rely fully on the implicit knowledge encoded in LMs for reasoning.
Probability-A (Pro-A) considers the generative probability of the choice conditioned on the question. However, it suffers from the statistical bias of choices, such as word frequency and sentence length~\citep{SensitivityofLanguageModels}. To alleviate this, MutualInfo-QA (MI-QA) calculates the mutual information between the question and the choice.
Another way to reduce the impact of statistical bias is to score each choice using the conditional probability of the question rather than the choice~\citep{ASimpleMethod,LMisAllYouNeedForCSReasoning} , which is denoted as Probability-Q (Pro-Q) in Table~\ref{tab:existing-methods}. 

\begin{table}[!t]
\centering
\small
\setlength{\extrarowheight}{0.2em}
\begin{tabular}{ll}\toprule[1pt]
Method & Score Function \\ \hline
Pro-A & $\left[P_{LM}(A|Q)\right]^{\frac{1}{|A|}}$         \\
Pro-Q & $\left[P_{LM}(Q|A)\right]^{\frac{1}{|Q|}}$         \\
MI-QA & $\left[\frac{P_{LM}(A|Q)}{P_{LM}(A)}\right]^{\frac{1}{|A|}}$ \\
\methodname (Ours) & $\sum_{S \in \mathbb{A}} \omega(S|A) P_{LM}(S|Q)$ \\
\bottomrule[1pt]
\end{tabular}
\caption{Three existing score functions and our method for unsupervised commonsense question answering. $Q$ is the question and $A$ is the choice.
$\mathbb{A}$ is the set of all possible answers and $\omega(S|A)$ is a weighting function defined in Eq.(\ref{eq:soft-indicator}). LM refers to a pre-trained language model, such as GPT-2 or BERT\protect\footnotemark~\citep{BERT}.}
\label{tab:existing-methods}
\vspace{-0.7em}
\end{table}
\footnotetext{$P_{BERT}(Q|A) \triangleq \prod_i^{|Q|} P_{BERT}(Q_{i}|Q_{/i},A)$. 
}

Some recent work claims that external knowledge can benefit commonsense reasoning.
Besides static knowledge bases (KBs), such as ConceptNet~\citep{ConceptNet} and Atomic~\citep{atomic}, there are also numerous studies treating LMs as dynamic KBs. \citet{LMasKB} shows that LMs can be used for KB completion. And \citet{KMfromLM} shows that BERT can distinguish true and fake ConceptNet triplets. 
Further, the extracted knowledge can work as complementary information for answering a question. 
\citet{ExplainYourself} proposes a model for CommonSenseQA~\citep{CommonsenseQA} that generates explanations for questions, which are then used as additional inputs. 
The shortcoming of this approach is that it requires collecting human explanations for each new dataset to fine-tune LMs. 
Some following researches explore unsupervised explanation/knowledge generator. CGA~\citep{dynamicReasoner} employs COMET~\citep{COMET} to generate intermediate inferences which are then used to score the choice. However, COMET is limited by a small set of question types so that CGA is difficult to generalize to different domains. Self-Talk~\citep{selfTalk} breaks the limit by extracting knowledge from GPT-2~\citep{GPT2}, which has no restriction on the query types. Thus, Self-Talk can be applied to a wide range of domains.
Despite the introduction of auxiliary information, these methods are essentially dependent on language model scores, so they are still sensitive to lexical perturbations.

Besides directly using pre-trained LMs, some recent efforts have been dedicated to automatically constructing task-specific data to train commonsense reasoners in zero-shot settings. 
\citet{UDSSM} and \citet{SurprisinglyRobustTrick} provide some rules to construct labeled training data from large corpus for pronoun disambiguation. 
\citet{KnowledgeTripletLearning}, \citet{Cosmo} and \citet{DataConstructionForZeroShotCQA} collect training data based on knowledge bases, such as Atomic~\citep{atomic}.
Though effective, they are limited by the specific task settings or highly dependent on the task-related knowledge bases, which makes them difficult to transfer to other commonsense reasoning tasks.

\section{Method}

In this paper, we focus on unsupervised multiple-choice commonsense question answering, which is formalized as follows: given a question and a set of choices, models should select the correct choice:
\par\nobreak
{
\begin{align} \nonumber
    \hat{A}=\argmax_{A} s(A|Q),
\end{align}
}%
where $s$ refers to a score function. Note that we have no access to any labeled task data.

\subsection{Motivation}
In existing unsupervised methods, the score functions are usually defined based on the language model scores. 
Taking Pro-A (Table~\ref{tab:existing-methods}) as an example, it first converts the question into a statement: 
\begin{itemize}[leftmargin=1em]
    \vspace{-0.3em}
    \item Q: I saw my breath when I exhaled. What was the cause of this? $\longrightarrow$ Rewrite: I saw my breath when I exhaled because \_\_\_
    \vspace{-0.3em}
\end{itemize}
And it then takes the statement as a prompt to calculate the generative probability of each choice. 
Note that the templates for rewriting is not the focus of this paper, and hence we directly use the templates of previous work~\citep{selfTalk,LMisAllYouNeedForCSReasoning} for our method and all the baselines in this paper (see Appendix for details).

Though successful, language model scores can be affected by many distracting factors, such as word frequency and sentence structure, etc. 
These factors can disturb the score functions to a large extent, as shown in Figure~\ref{fig:sensitive-examples}.
Our goal is to alleviate the influence of these distracting factors. Hence we propose a new method for unsupervised commonsense question answering, which achieves better results and performs more robustly.

\subsection{\methodname}
\label{sec:method-description}

\methodname is designed to predict the semantic score of an answer choice $A$. Instead of directly estimating the probability $P(A|Q)$ of the single choice $A$, the semantic score focuses on the probability $P(M_A|Q)$ where $M_A$ represents $A$'s semantics. Ideally, we decompose $P(M_A|Q)$ into the summation of the conditional probabilities of $A$'s \textit{supporters}, where the \textit{supporters} indicates all possible answers that have exactly the same semantics $M_A$.
%
%
Formally, the semantic score is defined as
%
\par\nobreak
{
\begin{align}
    s(A|Q) &\triangleq P(M_A|Q) = \sum_{S \in \mathbb{S}_A} P_{LM}(S|Q) \\
    &= \sum_{S \in \mathbb{A}} \mathbb{I}(S \in \mathbb{S}_A) P_{LM}(S|Q). \label{eq:hard-score}
\end{align}
}%
$\mathbb{S}_A$ is the set of \textit{supporters} of choice $A$, and $\mathbb{A}$ is the set of all possible answers. $\mathbb{I}(S \in \mathbb{S}_A)$ is an indicator function indicating whether $S$ is a \textit{supporter} of $A$. To obtain the \textit{supporter} set $\mathbb{S}_A$, we adopt a model to extract the sentence-level semantic features. Ideally, the indicator function is defined as
\par\nobreak
{
\begin{equation}
    \mathbb{I}(S \in \mathbb{S}_A) =
    \begin{cases}
        1 & \mbox{if } \cos(h_{S}, h_{A})=1, \\
        0 & \mbox{if } \cos(h_{S}, h_{A})<1,
    \end{cases} \label{eq:hard-indicator}
\end{equation}
}%
where $h_{A}$ is the semantic features of sentence $A$, and we assume that $S$ and $A$ are exactly the same in semantics if $h_{S}$ and $h_{A}$ point in the same direction.

However, Eq.(\ref{eq:hard-indicator}) uses a hard constraint that $\cos(h_{S}, h_{A})$ exactly equals to 1, which can be too strict to find acceptable \textit{supporters}. Therefore, we reformulate Eq.(\ref{eq:hard-score}) into a soft version:
\par\nobreak
{
\begin{align}
    s(A|Q) &\triangleq \sum_{S \in \mathbb{A}} \omega(S|A) P_{LM}(S|Q), \label{eq:soft-score}
\end{align}
}%
where the indicator function in Eq.(\ref{eq:hard-score}) is replaced by a soft function $\omega(S|A)$. To emulate $\mathbb{I}(S \in \mathbb{S}_A)$, $\omega(S|A)$ is expected to meet three requirements: (1) $\omega(S|A)\in [0,1]$ for any $S$ and $A$; (2) $\omega(S|A)=1$ if $\cos(h_{S}, h_{A})=1$; (3) $\omega(S|A)$ increases monotonically with $\cos(h_{S}, h_{A})$.
There are several different definitions of $\omega(S|A)$ meeting these requirements, which are explored in Section~\ref{sec:weight-function}. In this paper, $\omega(S|A)$ is defined as:
\par\nobreak
{
\begin{align}
    \omega(S|A) &= \frac{1}{Z(T)} \exp \left[ \frac{\cos(h_{S},h_{A})}{T} \right]. \label{eq:soft-indicator}
\end{align}
}%
$T$ is the temperature, and $Z(T)=\exp(\frac{1}{T})$ is a normalization term that makes $\omega(A|A)=1$. If $T \to 0$, $\omega(S|A)$ degenerates to the indicator function.
If $T > 0$, $\omega(S|A)$ relates to the von Mises-Fishers distribution over the unit sphere in the feature space, where the acceptable feature vectors are distributed around the mean direction $\frac{h_{A}}{||h_{A}||}$.

Since it is intractable to enumerate all possible answers in $\mathbb{A}$, we convert Eq.(\ref{eq:soft-score}) to an expectation over $P_{LM}(S|Q)$:
\par\nobreak
{
\small
\begin{align}
     s(A|Q) &= \mathbb{E}_{S \sim P_{LM}(S|Q)} \left[ \omega(S|A) \right] \notag \\
     & \approx \frac{1}{K} \sum_{i=1}^K \omega(S_i|A) \label{eq:voter-view} \\
     & = \frac{1}{K \cdot Z(T)} \sum_{i=1}^K \exp \left[ \frac{\cos(h_{S_i},h_{A})}{T} \right], \label{eq:modified-semantic-score}
\end{align}
}%
%
where $S_1,\cdots,S_K$ are sentences sampled from $P_{LM}(\cdot|Q)$, and $K$ is the sample size. 
$h_{A}$ and $h_{S_i}$ can be extracted from a pre-trained model, e.g., SentenceBERT \citep{sentenceBERT}.

From Eq.(\ref{eq:modified-semantic-score}), we can see the semantic score $s(A|Q)$ is only dependent on the semantic feature $h_A$ and regardless of $A$'s surface form.
Therefore, our method will produce similar semantic scores for synonymous choices, assuming that the synonymous choices have similar semantic features.


\begin{figure}[ht]
    \centering
    \includegraphics[width=0.95\linewidth]{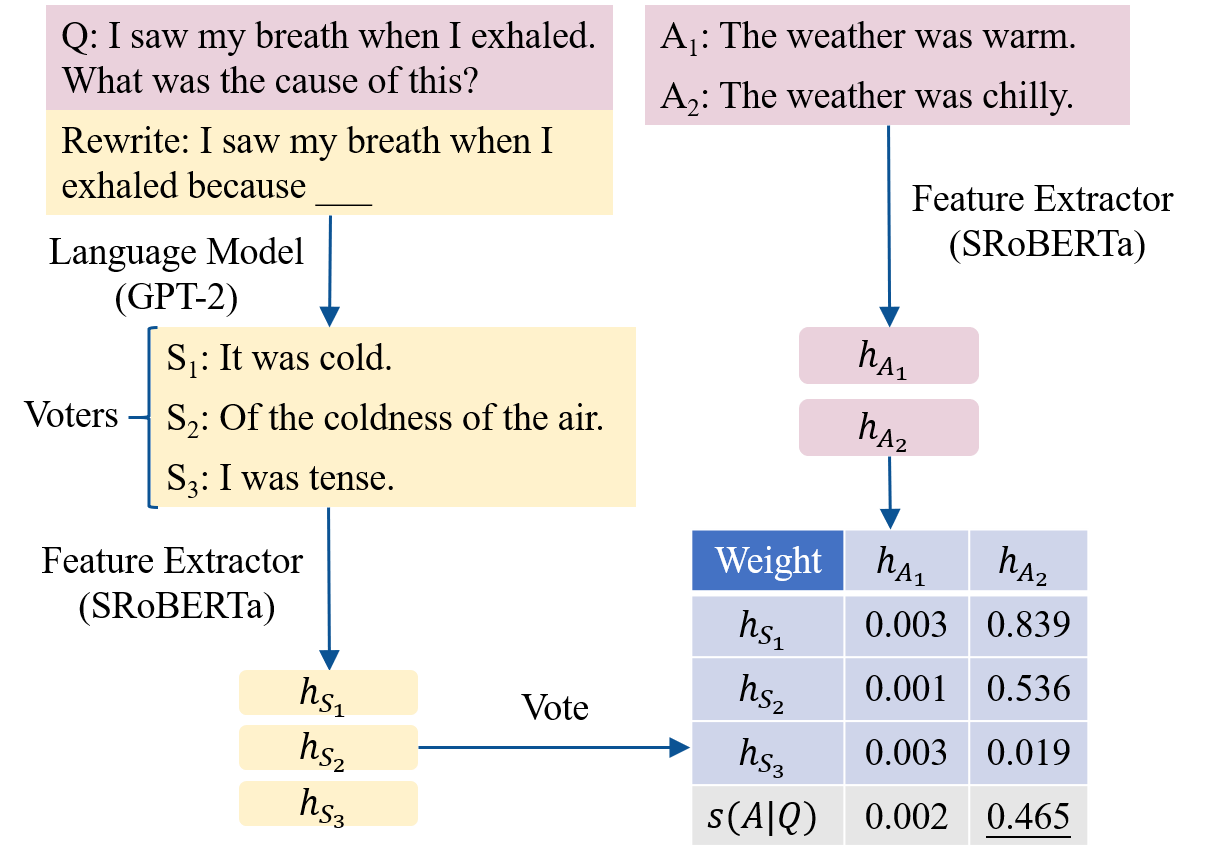}
    \caption{Process of \methodname in the view of voting. We use the same templates with previous work~\citep{selfTalk,LMisAllYouNeedForCSReasoning} to rewrite interrogative sentences into declarative ones. And then use GPT-2 to generate some plausible answers as \textit{voters} $S_i$, conditioned on the rewritten question. The choices and \textit{voters} are encoded via SentenceRoBERTa to obtain semantic features, $h_{A_j}$ and $h_{S_i}$, which are then used to calculate the voting weights $\omega(S_i|A_j)$. The choice with the largest score $s(A_j|Q)$ is selected as the answer.}
    \label{fig:method-overview}
    \vspace{-1.5em}
\end{figure}

\subsection{The Voting View of \methodname}

\label{sec:another-view}


At the beginning of Section \ref{sec:method-description}, we define the semantic score as the summation of the conditional probabilities over the \textit{supporters}. However, in Eq.(\ref{eq:modified-semantic-score}), the sampled sentences $S_1, \cdots, S_K$ are not $A$'s \textit{supporters} because they may not be semantically similar to $A$. To address the differences, we name the sampled sentences $S_1, \cdots, S_K$ as \textit{voters}, which are plausible answers to the question $Q$. 
In this section, we will show another view of our method, which works like a procedure that the \textit{voters} vote out the correct choice.

Suppose there are two candidate choices $A_1$ and $A_2$, our method is to find the correct choice according to the semantic scores, $s(A_1|Q)$ and $s(A_2|Q)$.
Following Eq.(\ref{eq:voter-view}), our method can be decomposed into two steps:
First, \textbf{sample} some \textit{voters} $S_1, \cdots, S_K$ from $P_{LM}(\cdot|Q)$. This step only considers the question $Q$ but no candidate choices.
Second, each \textit{voter} votes for the choices with the semantic similarity weights. For example, $S_i$ votes for $A_j$ with the weight of $\omega(S_i|A_j)$.
%
%
The candidate choice that receives more votes will have a higher semantic score and be selected as the final answer.

Figure~\ref{fig:method-overview} shows the process of \methodname in the view of voting. 
Although the voting view is intuitive, the formalism in Section \ref{sec:method-description} provides more insights: 
(1) Our method approximates the probability of semantics, which works as the theoretical basis of \methodname. 
(2) Our method can be seen as an extension of Pro-A (see Table \ref{tab:existing-methods}), since Pro-A only calculates the language model score for a single sentence, whereas our method calculates the semantic score for a set of \textit{supporters}.
(3) Eq.(\ref{eq:soft-score}) provides guidance, the three requirements mention before, for the design of the voting weight function $\omega(S|A)$.
Specifically, the guidance explains the rationality of the formulation of Eq.(\ref{eq:soft-indicator}).




\section{Experiments}

\subsection{Datasets}

We conducted experiments on four multiple-choice commonsense question answering tasks, COPA~\citep{COPA}, StoryClozeTest (SCT)~\citep{StoryClozeTest}, SocialIQA~\citep{SocialIQA} and CosmosQA~\citep{CosmosQA}. 
For each instance, only one choice is correct. 
See Appendix for more description about datasets.





For COPA, we reported the results on its test set.
As the test sets of another three datasets are hidden, for convenience of analysis,
we reported the experiment results on their development sets.


\subsection{Baselines}

We employed five strong baselines.
Table~\ref{tab:existing-methods} shows three of them, \textbf{Pro-A}, \textbf{Pro-Q} and \textbf{MI-QA}.
%
There is no explicit auxiliary information used in these three methods, while another two baselines rely on explicit information supplementation.
\textbf{CGA}~\citep{dynamicReasoner} and \textbf{Self-Talk}~\citep{selfTalk} query pre-trained language models (e.g., GPT-2, COMET~\citep{COMET}) for relevant knowledge, which forms part of contexts. And then, similar to Pro-A, they take the generative probabilities of choices as scores.

\begin{table*}[ht]
\centering
\small
\adjustbox{max width=\textwidth}{
\begin{tabular}{lllccccc}\toprule[1pt]
Dataset                    & Method            & \multicolumn{1}{l}{\begin{tabular}[c]{@{}l@{}}Pre-trained\\ Models\end{tabular}} & \multicolumn{1}{l}{\begin{tabular}[c]{@{}l@{}}Original\\ Accuracy ($\uparrow$)\end{tabular}} & \multicolumn{1}{l}{\begin{tabular}[c]{@{}l@{}}After-Attack\\ Accuracy ($\uparrow$)\end{tabular}} & \multicolumn{1}{l}{\begin{tabular}[c]{@{}l@{}}Attack \\ Success Rate ($\downarrow$)\end{tabular}} & \multicolumn{1}{l}{\begin{tabular}[c]{@{}l@{}}Percentage of\\ Perturbed Words\end{tabular}} & \multicolumn{1}{l}{\begin{tabular}[c]{@{}l@{}}Semantic\\ Similarity\end{tabular}} \\ \hline
\multirow{6}{*}{COPA}      & Pro-A             & GPT-2 & 73.6                                                        & 4.6                                                             & 93.8                                                           & 17.3               & 0.883                                                         \\
                           & Pro-Q             & RoBERTa & \textbf{79.4}                                                        & 23.0                                                            & 71.0                                                           & 22.9               & 0.828                                                         \\
                           & MI-QA             & GPT-2 & 74.6                                                        & 16.2                                                            & 78.3                                                           & 19.9               & 0.865                                                         \\
                          & Self-talk         & COMET+GPT-2 & 68.6                                                        & 8.4                                                             & 87.8                                                          & 19.8               & 0.855                                                          \\
                          & CGA  & GPT-2 & 72.2                                                        & 4.8                                                            & 93.4                                                          & 17.1              &  0.886                                                        \\
                           & \bf \methodname & GPT-2+SRoBERTa & \textbf{79.4}                                                        & \textbf{59.0}                                                            & \textbf{25.7}                                                           & 21.7               & 0.827
                           \\ \hline
\multirow{6}{*}{SCT}       & Pro-A             & GPT-2 & 72.3                                                        & 4.8                                                             & 93.3                                                           & 14.3               & 0.917                                                         \\
                           & Pro-Q             & RoBERTa & 56.3                                                        & 22.3                                                            & 60.3                                                           & 18.1               & 0.872                                                         \\
                           & MI-QA             & GPT-2 & 66.1                                                        & 29.2                                                            & 55.8                                                           & 16.2               & 0.885                                                         \\
                          & Self-talk         & COMET+GPT-2 & 70.4                                                        & 4.7                                                             & 93.3                                                           & 14.2                & 0.915                                                         \\
                          & CGA  & GPT-2 & 71.5                                                       & 4.8                                                             & 93.2                                                          & 14.3               & 0.916                                                         \\
                           & \bf \methodname & GPT-2+SRoBERTa & \textbf{83.2}                                                        & \textbf{69.4}                                                            & \textbf{16.5}                                                           & 18.3               & 0.856
                           \\ \hline
\multirow{6}{*}{SocialIQA} & Pro-A             & GPT-2 & 46.0                                                        & 16.2                                                            & 64.7                                                           & 21.0               & 0.876                                                         \\
                           & Pro-Q             & RoBERTa & 42.2                                                        & 27.8                                                            & 34.2                                                           & 23.2               & 0.843                                                         \\
                           & MI-QA             & GPT-2 & 41.2                                                        & 24.6                                                            & 40.4                                                           & 25.3               & 0.866                                                         \\
                          & Self-talk         & COMET+GPT-2 & \textbf{47.5}                                                       & 12.3                                                            & 74.0                                                           & 22.2               & 0.872                                                         \\
                          & CGA  & COMET & 45.4                                                        & 18.4                                                           & 59.4                                                           & 22.3               & 0.867                                                         \\
                           & \bf \methodname & GPT-2+SRoBERTa & \textbf{47.5}                                                        & \textbf{38.2}                                                            & \textbf{19.5}                                                           & 23.5               & 0.839
                           \\ \hline
\multirow{6}{*}{CosmosQA}  & Pro-A             & GPT-2 & 36.8                                                        & 1.3                                                             & 96.4                                                           & 9.2                & 0.927                                                         \\
                           & Pro-Q             & RoBERTa & 21.5                                                        & 5.0                                                             & 76.6                                                           & 13.7               & 0.859                                                         \\
                           & MI-QA             & GPT-2 & 29.3                                                        & 7.4                                                             & 74.8                                                           & 12.1               & 0.886                                                         \\
                          & Self-talk         & COMET+GPT-2 & 36.1                                                        & 1.2                                                             & 96.7                                                           & 8.9                & 0.928                                                         \\
                          & CGA  & GPT-2 & 42.4                                                        & 1.7                                                             & 96.0                                                           & 9.6               & 0.924                                                         \\
                           & \bf \methodname & GPT-2+SRoBERTa & \textbf{56.1}                                                        & \textbf{32.6}                                                            & \textbf{41.8}                                                           & 13.9               & 0.859
\\\bottomrule[1pt]
\end{tabular}}
\caption{Evaluation results, including the original selection accuracy before attack, the accuracy after attack, the attack success rate, the percentage of perturbed words with respect to the original sentence length in successful attacks, and the semantic similarity between the original and paraphrased choices. GPT-2, RoBERTa and SRoBERTa refer to GPT-2-xlarge, RoBERTa-large~\citep{roberta} and SentenceRoBERTa-large, respectively.}
\label{tab:main-results}
\vspace{-1.2em}
\end{table*}

\subsection{Experiment Settings}

For each method, we tried different pre-trained language models (see Appendix for details), and then selected the pre-trained LMs that maximized the accuracy on each dataset.
The details of the selection of pre-trained LMs can be found in Table~\ref{tab:main-results}.

For \methodname, we used GPT-2 to generate \textit{voters} via Nucleus Sampling~\citep{NucleusSampling} with $p=0.9$. 
The sample size $K$ of \textit{voters} is set to $500$. 
In Section~\ref{sec:ablation-sample-number}, we show that a small sample size can also lead to superior performance.
Self-Talk and CGA also rely on the generated answers from GPT-2 or COMET. Different from \methodname, for these two baselines, more generated answers will not always lead to better performance (see Section~\ref{sec:ablation-sample-number}). Thus, we selected the optimal sample size for them rather than the same sample size with \methodname.

When evaluating \methodname on COPA, we tuned the temperature $T$ on its development set, and then reported the results on the test set with the tuned temperature $T=0.1$. Due to the absence of test sets of other datasets, we evaluated \methodname on their development sets without tuning the temperature and directly set $T=0.1$.

\subsection{Main Results}

Table~\ref{tab:main-results} shows the evaluation results about accuracy and robustness.

\subsubsection{Accuracy}
Among all the methods, \methodname achieved the best performance on all the datasets.
Especially on SCT and CosmosQA, \methodname outperformed the best baselines by more than 10 points. 
It can be inferred that the semantic scores are beneficial for commonsense question answering due to the reduction of distracting factors.
Pro-Q performed better than other baselines on COPA, perhaps because it suffered less from the statistic bias of choices~\citep{LMisAllYouNeedForCSReasoning}. 
However, Pro-Q lost its superiority on another three datasets, because it is unsuitable for processing long or complex contexts.

\subsubsection{Robustness}

To test the robustness under the synonym replacement attack, we used TextFooler~\citep{TextFooler} to attack the methods
by perturbing the correct choices of the correctly predicted examples.
The percentage of perturbed words refers to what percentage of words in choices are replaced
in successful attacks. The semantic similarity is measured between the paraphrased choice and the original choice.
Considering the attack success rate and the after-attack accuracy, \methodname is much more robust than all baselines. To be specific, the attack success rates on \methodname are at least $39$ points lower than those of Pro-A, CGA, and Self-Talk on all datasets.
MI-QA and Pro-Q are designed to reduce the impact of statistic bias in choices, so that they can resist lexical perturbation 
to some extent.
Even so, \methodname is remarkably lower than MI-QA and Pro-Q in terms of attack success rates on all datasets. 

An observation is that the attack success rate on \methodname on CosmosQA is higher than those on the other datasets. The reason is that, the contexts in CosmosQA are so complex that GPT-2 is more difficult to generate high-quality answers. If there is a more powerful generator, the robustness of \methodname is expected to have a further improvement.

\subsection{Consistency Testing}

We have claimed that a commonsense question answering method should assign close scores to synonymous choices. To verify that \methodname better meets this requirement, we conducted consistency testing for all the methods on four datasets.
For each example, the consistency testing of a method is conducted in three steps: 
(1) Originally, the example has one correct and several wrong answer choices. We randomly sample some choices from other examples as additional wrong choices. After that, the example will have one correct choice and 19 wrong choices.
(2) Leverage a commonly used automatic translation service, Baidu Translation, to translate each choice from English into an intermediate language, and then back-translate it into English. During this process, we employ three intermediate languages, Chinese, Spanish, and Russian, because the translation quality of these languages is better than others. As a result, each choice is accompanied with three synonymous choices. 
(3) Use the commonsense question answering method to calculate the scores for each choice as well as its synonymous choices, and then sort all the choices according to their scores. 
Because the scoring scales of these methods are different, 
we calculate the standard deviation of the ranks of the correct choice and its synonymous choices.

\begin{table}[]
\centering
\small
\scalebox{0.87}{
\begin{tabular}{lcccc}\toprule[1pt]
Method / Dataset  & COPA & SCT & SocialIQA & CosmosQA \\ \hline
Pro-A             & 9.1  &11.0 & 11.7      &  9.4    \\
Pro-Q             & 6.9  &8.5  & 11.6      & 12.3 \\
MI-QA             & 7.5  & 5.8 & 11.1      & 7.9  \\ \hline
Self-Talk         & 13.3 & 9.5 & 10.7      & 10.1 \\
CGA  & 9.7  & 11.0 & 10.9      & 9.5  \\ \hline
\bf \methodname & \textbf{4.1}  & \textbf{3.2} & \textbf{5.8}       &  \textbf{4.7} 
\\\bottomrule[1pt]
\end{tabular}}
\caption{Consistency testing where the methods rank 80 choices to find 4 correct ones for each example. The metric is the standard deviation of the ranks of 4 correct synonymous choices averaged over 500 examples.}
\label{tab:consistency-testing}
\vspace{-0.7em}
\end{table}

Table~\ref{tab:consistency-testing} shows the average standard deviation of the ranks. As expected, the average standard deviation of \methodname is much lower than any other method on all the datasets, confirming that \methodname assigns more similar ranks and closer scores to synonymous choices.
We also observed that MI-QA provided relatively stable predictions compared with other baseline methods. A possible explanation is that, the normalization term $P_{LM}(A)$ helps alleviate the influence of lexical perturbations.

\subsection{Trends of Accuracy with Answer Length}

Answer length is also a type of distracting factor which may mislead baseline methods.
To explore to which extent answer lengths affect the performance of methods, we divided the development set of CosmosQA into four subsets according to the length of correct choice. Table~\ref{tab:trends-accuracy-length} shows the results of \methodname and a robust baseline, MI-QA.
Compared with MI-QA, \methodname has much more stable performance as answer lengths vary. The reason is that, \methodname focuses on semantic information so that it has stronger resistance to such distracting factors.

\begin{table}[]
\centering
\scalebox{0.87}{
\begin{tabular}{lccccc}\toprule[1pt]
\multicolumn{1}{c}{\multirow{2}{*}{Method}} & \multicolumn{5}{c}{Answer Length}  \\
\multicolumn{1}{c}{}                        & All  & [1,5]  & [6,10] & [11,15] & [16,20] \\\hline
MI-QA                                       & 29.3 & 51.6 & 27.9 & 24.4  & 23.8  \\
\methodname                                 & 56.1 & 58.6 & 58.0 & 54.1  & 51.2  \\\bottomrule[1pt]
\end{tabular}}
\caption{The trends of accuracy with answer length for \methodname and MI-QA on CosmosQA.}
\label{tab:trends-accuracy-length}
\end{table}

\subsection{Ablation Study}

\subsubsection{Analysis on Temperature}\label{sec:ablation-temperature}

In the previous experiments, the temperature $T$ of \methodname was set to $0.1$ by default.
To investigate the influence of $T$, we varied $T$ in a wide range from $0.05$ to $10$ and report the results in Table~\ref{tab:ablation-temperature}. 
Considering that the temperature varied greatly, the performance of \methodname is relatively stable, indicating that \methodname is not so sensitive to the selection of $T$. 
Another observation is that,
although the four datasets are different in domains and text length, the trends of performance with temperature on them are relatively similar, illustrating that the temperature selected on one task can be generalized to other tasks.

\begin{table}[t]
\small
\centering
\scalebox{0.87}{
\begin{tabular}{lclclclcl}\toprule[1pt]
\multirow{2}{*}{$T$} & \multicolumn{2}{c}{COPA}       & \multicolumn{2}{c}{SCT}        & \multicolumn{2}{c}{SocialIQA}  & \multicolumn{2}{c}{CosmosQA}   \\
                                       & Bef  & \multicolumn{1}{c}{Aft} & Bef  & \multicolumn{1}{c}{Aft} & Bef  & \multicolumn{1}{c}{Aft} & Bef  & \multicolumn{1}{c}{Aft} \\ \hline
10                                    & 75.6 & 48.8                    & 82.0 & 64.7                    & 46.3 & 35.9                    & 52.7 & 22.3                    \\
1                                      & 76.4 & 48.8                    & 82.4 & 64.5                    & 46.6 & 36.1                    & 53.3 & 22.4                    \\
0.2                                      & 77.0 & 52.8                    & \textbf{83.6} & 66.3                    & 46.9 & 36.8                    & 54.8 & 26.1                     \\
0.1                                     & 79.4 & \textbf{59.0}                    & 83.2 & \textbf{69.4}                    & \textbf{47.5} & \textbf{38.2}                    & \textbf{56.1} & \textbf{32.6}                    \\
0.05                                     & \textbf{80.2} & 54.6                    & 80.8 & 61.4                    & 46.0 & 36.5                    & 55.1 & 28.8                    \\
\bottomrule[1pt]
\end{tabular}}
\caption{The before-attack (Bef) and after-attack (Aft) accuracy of \methodname with different temperatures.}
\label{tab:ablation-temperature}
\vspace{-0.5em}
\end{table}

\subsubsection{Analysis on Sample Size}\label{sec:ablation-sample-number}

\begin{figure}[t]
    \centering
    \includegraphics[width=\linewidth]{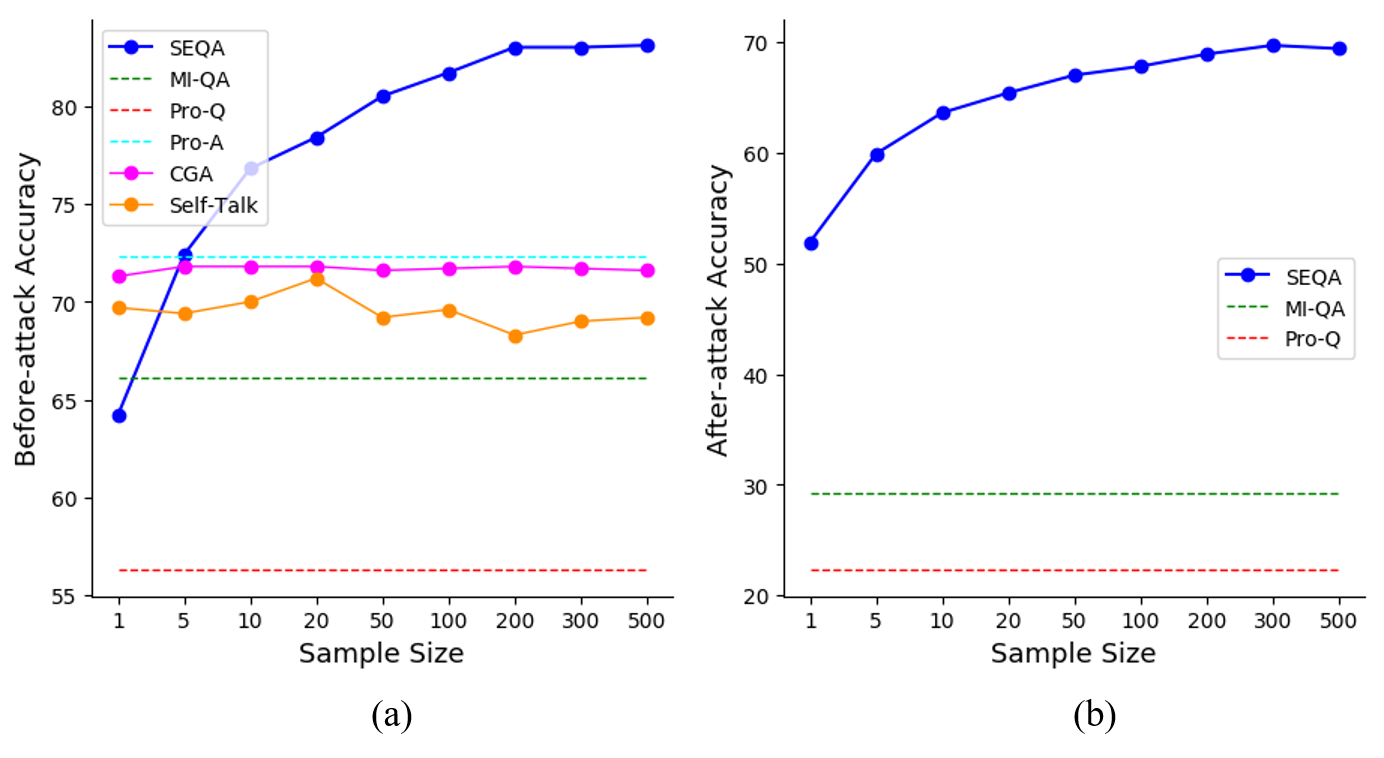}
    \caption{The before-attack (a) and after-attack accuracy (b) of methods with different sample sizes on SCT. The after-attack accuracy of Pro-A, CGA and Self-Talk is below $5.0\%$, and thus omitted in (b).}
    \label{fig:ablation-sample-number}
    \vspace{-0.5em}
\end{figure}

Figure~\ref{fig:ablation-sample-number} shows the effect of the sample size $K$ on \methodname. 
For comparison, Figure~\ref{fig:ablation-sample-number} also includes the results of baselines in the settings of before- and after-attack, respectively.
Due to the limitation of space, the results on the other datasets are shown in Appendix.
As expected, the before-attack and after-attack accuracy on SCT increased with the sample size.
In detail, the rapid increase in performance occurred when $K<100$, and then the improvement slowed down when $K>100$. Finally, \methodname achieved a stable and relatively high performance.

CGA and Self-Talk also leverage LMs to generate some plausible answers. Different from our method, they use the generated answers to form part of the question, and then calculate the generative probability of the choice based on the augmented question. We also tried different sample sizes for the two methods, and Figure~\ref{fig:ablation-sample-number} (a) shows that their accuracy will not stably increase with a larger sample size.


\subsubsection{Analysis on $\omega(S|A)$}\label{sec:weight-function}

$\omega(S|A)$ in \methodname can be defined in different forms, as long as the three requirements mentioned in Section~\ref{sec:method-description} are met.
Besides the default definition, we explored another three forms of $\omega(S|A)$, and the experiment results on COPA are shown in Table~\ref{tab:weight-function-selection}.
Although the performance varies with $\omega(S|A)$, the before-attack accuracy of \methodname still outperformed most of the baselines with any definition of $\omega(S|A)$. Moreover, \methodname maintains its obvious advantage in after-attack accuracy, which reflects the inherent robustness of \methodname.

\begin{table}[]
\small
\centering
\begin{tabular}{lcc}\toprule[1pt]
$\omega(S|A)=\frac{1}{f(1)}f\left(\cos(h_S,h_A)\right)$                                                   & Bef  & Aft  \\ \hline
$f(x) = \mathbb{I}(x>\alpha)$                          & 77.2 & 47.2     \\
$f(x) = {\rm ReLU}(x-\beta)$                          & 77.6 & 45.2     \\
$f(x) = {\rm sigmoid}(\frac{x}{T})$                              & 75.6 & 48.6    \\
$f(x) = \exp \left( \frac{x}{T} \right)$ & 79.4 & 59.0 
\\\bottomrule[1pt]
\end{tabular}
\caption{The before-attack (Bef) and after-attack (Aft) accuracy of \methodname on the test set of COPA with different definitions of $\omega(S|A)$. $\alpha, \beta, T_1, T_2$ are hyperparameters tuned on the development set of COPA.}
\label{tab:weight-function-selection}
\vspace{-0.5em}
\end{table}

\begin{table}[]
\small
\centering
\begin{tabular}{lccc}\toprule[1pt]
 & \multicolumn{3}{c}{GPT-2} \\
   & medium  & large & xlarge \\ \hline
Avg. GloVe                           & 56.6    & 59.6  & 61.2   \\
SBERT-base                           & 71.2    & 72.6  & 74.8    \\
SRoBERTa-base                        & 72.4    & 72.0  & 75.4    \\
SRoBERTa-large                       & 74.2    & 75.2  & 79.4   
\\\bottomrule[1pt]
\end{tabular}
\caption{\methodname's accuracy with different feature extractors and language models on COPA. Avg. GloVe means the average pooling of the pre-trained word embeddings \cite{GloVe} over the sentence.  
}
\label{tab:generator-encoder-selection}
\vspace{-0.5em}
\end{table}

\subsubsection{Analysis on Pre-trained Language Model and Feature Extractor}\label{sec:analysis_generator_encoder}

\methodname has no limit on the selection of the pre-trained language model and the feature extractor. Table~\ref{tab:generator-encoder-selection} shows how the accuracy of \methodname on COPA varied with the language model and the feature extractor. 
As expected, more powerful extractor usually led to higher accuracy under the same settings of language models. Similar conclusion can be obtained for the language model.
It can be inferred that, if there are more powerful language models or feature extractors in the future, the performance of \methodname may be further improved.

\subsection{Analysis on the Quality of \textit{Voters}}\label{sec:human-evaluation}


While the performance of \methodname served as an extrinsic evaluation for the quality of the \textit{voters} (plausible answers sampled from $P_{LM}(\cdot|Q)$, described in Section \ref{sec:another-view}), we were also interested in evaluating it intrinsically.
We sampled 125 \textit{voters} from COPA. For each \textit{voter}, we provided crowd-sourcing workers with the original question, and asked them: 1) whether the \textit{voter} is grammatical, not entirely grammatical but understandable, or completely not understandable, 2) whether the \textit{voter} is a reasonable answer to the question, not reasonable but relevant, or completely irrelevant. 
These evaluation tasks comprehensively examined the \textit{voters} in grammar and logicality.
The annotation tasks were carried out in Amazon Mechanical Turk, and we aggregated annotations from 3 workers using majority vote. 

\begin{table}[t]
\small
\centering
\begin{tabular}{lccc}\toprule[1pt]
Score      & 3      & 2      & 1      \\ \hline
Grammar   & 84.8\% & 12.8\% & 2.4\% \\
Logic    & 40.8\% & 25.6\% & 33.6\% \\\bottomrule[1pt]
\end{tabular}
\caption{Manual evaluation of the quality of \textit{voters} (generated by GPT-2-xlarge conditioned on questions). Score 3/2/1 correspond to high, middle and low quality, respectively, in terms of grammar and logicality.}
\label{tab:human-evaluation-generated-answers}
\vspace{-0.5em}
\end{table}

Table~\ref{tab:human-evaluation-generated-answers} shows the results of the human evaluation of the \textit{voters}. Score 3/2/1 correspond to the high, middle and low quality, respectively.
According to the grammar scores, $97.6\%$ of the \textit{voters} are grammatical or at least understandable, for which most of the \textit{voters} belong to the natural language space.
In terms of logicality, $40.8\%$ of the \textit{voters} are reasonable answers to the questions, which may not be very satisfying.
However, in Section~\ref{sec:voting-weight-distribution}, we will show that \methodname makes prediction based on a small part of \textit{voters}, and hence \methodname is robust even though there are some irrelevant \textit{voters}.



\subsection{Voting Weight Distribution}\label{sec:voting-weight-distribution}

\begin{figure}[t]
    \centering
    \includegraphics[width=0.9\linewidth]{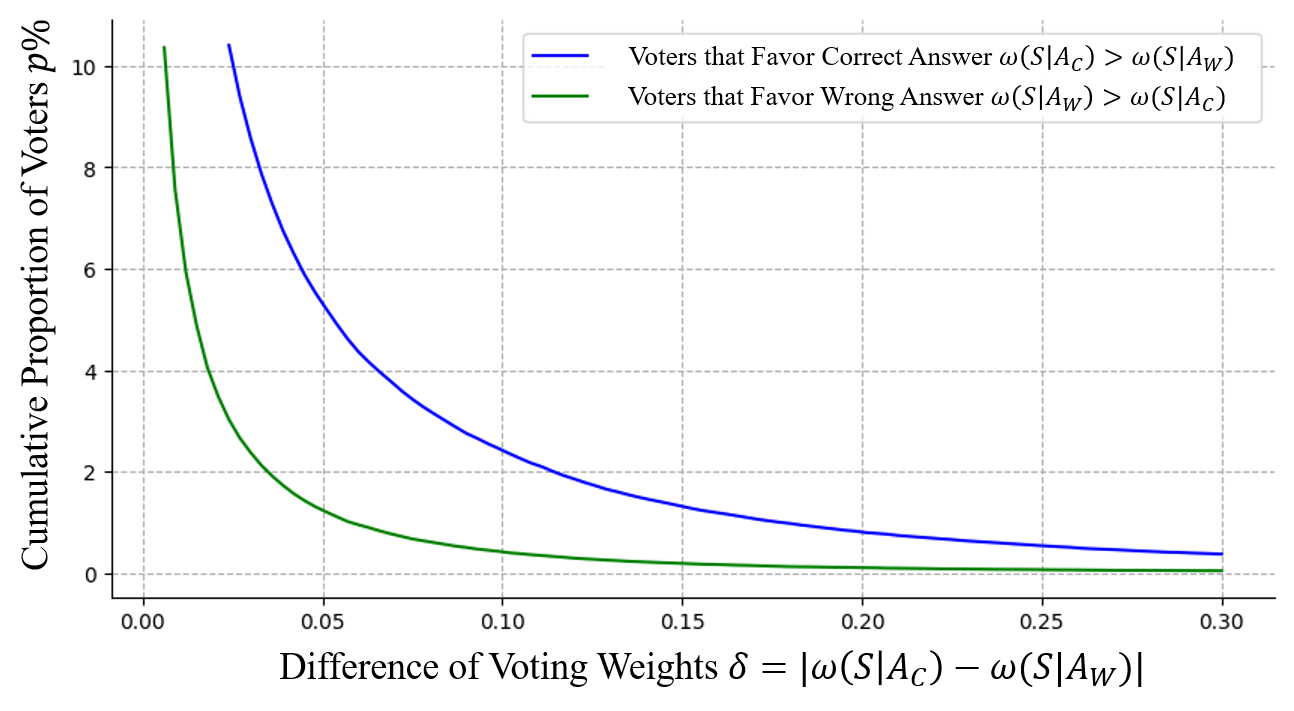}
    \caption{The cumulative proportion of \textit{voters} favoring the correct answer $A_C$ or the wrong answer $A_W$ on COPA. 
    Each point $(\delta, p)$ means that $p\%$ of \textit{voters} satisfy $|\omega(S|A_C)-\omega(S|A_W)| \geq \delta$, where $S$ refers to a \textit{voter}. The area between the two curves equals to the difference of the semantic scores $s(A_C|Q) - s(A_W|Q)$.}
    \label{fig:score-distribution}
    \vspace{-0.5em}
\end{figure}

We visualize the cumulative proportion of \textit{voters} favoring the correct or the wrong choices (see Figure~\ref{fig:score-distribution}). The curve is averaged over all instances in the test set of COPA, where we sampled 500 \textit{voters} for each instance and set $T=0.1$.

From the curves, we can find several properties of \textit{voters}:
(1) The \textit{voters} favor the correct choices over the wrong choices, where the curve for correct choices is consistently above the curve for wrong ones. The area between two curves shows the difference of semantic scores $s(A_C|Q) - s(A_W|Q)$, which is a large gap compared with the area under the bottom curve.
%
(2) $93.5\%$ of \textit{voters} do not strongly favor any choices ($|\omega(S|{A_C})-\omega(S|{A_W})|< 0.05$), indicating that they are semantically irrelevant to both candidate choices. However, Table \ref{tab:human-evaluation-generated-answers} shows that $40.8\%$ of \textit{voters} are logically reasonable, so many \textit{voters} are reasonable but irrelevant to both answers. It suggests that there can be several reasonable answers for a single question, and the sampled \textit{voters} are diverse in the semantics.
(3) Although there are only $5.3\%$ of \textit{voters} strongly favoring the correct choices, there are much less \textit{voters} ($1.2\%$) favoring the wrong ones. It explains why our method is able to predict the correct answer.



\begin{table}[t]
\centering
\small
\begin{tabular}{cp{4cm}c}\toprule[1pt]
\multicolumn{3}{l}{Q: The car ran out of gas. What happened as a result?}                        \\
\multicolumn{3}{l}{$A_C$: The driver was stranded on the road. (\cmark)}                                     \\
\multicolumn{3}{l}{$A_W$: The driver picked up a hitchhiker. (\xmark)}                                       \\ \hline
\multirow{2}{*}{$\omega(S_i|A_C)$} & \multicolumn{1}{c}{\multirow{2}{*}{\textit{voter}}}               & \multirow{2}{*}{$\omega(S_i|A_W)$} \\
                   & \multicolumn{1}{c}{}                                     &                    \\
\hline
0.161             & I had to park on a dead end road.                        & 0.008             \\
0.008             & We picked up a hitchhiker and she drove us to the diner. & 0.137             \\
0.013             & We stopped at a gas station.                             & 0.011             \\
0.018             & It was time to hit the road again.                       & 0.010      
\\\bottomrule[1pt]
\end{tabular}
\caption{An example of \textit{voters} as well as their voting weights. $A_C$ is the correct choice, while $A_W$ is wrong. $S_i$ refers to a \textit{voter}.}
\label{tab:semantic-reasoner-case}
\vspace{-0.5em}
\end{table}

To help understand the relationship between \textit{voters} and choices,
Table~\ref{tab:semantic-reasoner-case} provides an instance with \textit{voters} and their voting weights to the choices. 
We show four types of \textit{voters}: favoring the correct choice, favoring the wrong choice, logically reasonable but not favoring either choices, and unreasonable and irrelevant to both choices.
We can see that the last two types of \textit{voters} can hardly affect the method's prediction, because their voting weights are much smaller than the first two types of \textit{voters}.
%


\section{Conclusion} 

We present a semantic-based question answering method, \methodname, which can answer commonsense questions more accurately and robustly in an unsupervised setting. Instead of directly scoring each answer choice, our method focuses on the probability of observing a choice's semantics.
In the view of voting, \methodname first generates some plausible answers (\textit{voters}) and then utilizes them to vote for the correct choice by considering the semantic similarity between each choice and each \textit{voter}. 
Experiment results show that \methodname achieves the best performance on four datasets, and it is remarkably more robust than all the baselines when being attacked by TextFooler. 


\section*{Acknowledgments}
This work was partly supported by the NSFC projects (Key project with No. 61936010 and regular project with No. 61876096). 
This work was also supported by the Guoqiang Institute of Tsinghua University, with Grant No. 2019GQG1 and 2020GQG0005.
This work was also supported by Huawei Noah's Ark Lab.

\bibliographystyle{acl_natbib}
\bibliography{anthology,acl2021}

\begin{thebibliography}{31}
\expandafter\ifx\csname natexlab\endcsname\relax\def\natexlab#1{#1}\fi

\bibitem[{Abdou et~al.(2020)Abdou, Ravishankar, Barrett, Belinkov, Elliott, and
  S{\o}gaard}]{SensitivityofLanguageModels}
Mostafa Abdou, Vinit Ravishankar, Maria Barrett, Yonatan Belinkov, Desmond
  Elliott, and Anders S{\o}gaard. 2020.
\newblock \href {https://www.aclweb.org/anthology/2020.acl-main.679/} {The
  sensitivity of language models and humans to winograd schema perturbations}.
\newblock In \emph{ACL}, pages 7590--7604.

\bibitem[{Banerjee and Baral(2020)}]{KnowledgeTripletLearning}
Pratyay Banerjee and Chitta Baral. 2020.
\newblock \href {https://arxiv.org/abs/2005.00316} {Self-supervised knowledge
  triplet learning for zero-shot question answering}.
\newblock \emph{CoRR}.

\bibitem[{Bosselut and Choi(2019)}]{dynamicReasoner}
Antoine Bosselut and Yejin Choi. 2019.
\newblock \href {http://arxiv.org/abs/1911.03876} {Dynamic knowledge graph
  construction for zero-shot commonsense question answering}.
\newblock \emph{CoRR}.

\bibitem[{Bosselut et~al.(2019)Bosselut, Rashkin, Sap, Malaviya,
  {\c{C}}elikyilmaz, and Choi}]{COMET}
Antoine Bosselut, Hannah Rashkin, Maarten Sap, Chaitanya Malaviya, Asli
  {\c{C}}elikyilmaz, and Yejin Choi. 2019.
\newblock \href {https://doi.org/10.18653/v1/p19-1470} {{COMET:} commonsense
  transformers for automatic knowledge graph construction}.
\newblock In \emph{ACL}, pages 4762--4779.

\bibitem[{Davison et~al.(2019)Davison, Feldman, and Rush}]{KMfromLM}
Joe Davison, Joshua Feldman, and Alexander~M. Rush. 2019.
\newblock \href {https://doi.org/10.18653/v1/D19-1109} {Commonsense knowledge
  mining from pretrained models}.
\newblock In \emph{EMNLP-IJCNLP}, pages 1173--1178.

\bibitem[{Devlin et~al.(2019)Devlin, Chang, Lee, and Toutanova}]{BERT}
Jacob Devlin, Ming{-}Wei Chang, Kenton Lee, and Kristina Toutanova. 2019.
\newblock \href {https://doi.org/10.18653/v1/n19-1423} {{BERT:} pre-training of
  deep bidirectional transformers for language understanding}.
\newblock In \emph{NAACL-HLT}, pages 4171--4186.

\bibitem[{Holtzman et~al.(2020)Holtzman, Buys, Du, Forbes, and
  Choi}]{NucleusSampling}
Ari Holtzman, Jan Buys, Li~Du, Maxwell Forbes, and Yejin Choi. 2020.
\newblock \href {https://openreview.net/forum?id=rygGQyrFvH} {The curious case
  of neural text degeneration}.
\newblock In \emph{ICLR}.

\bibitem[{Huang et~al.(2019)Huang, Bras, Bhagavatula, and Choi}]{CosmosQA}
Lifu Huang, Ronan~Le Bras, Chandra Bhagavatula, and Yejin Choi. 2019.
\newblock \href {https://doi.org/10.18653/v1/D19-1243} {Cosmos {QA:} machine
  reading comprehension with contextual commonsense reasoning}.
\newblock In \emph{EMNLP}, pages 2391--2401.

\bibitem[{Jin et~al.(2020)Jin, Jin, Zhou, and Szolovits}]{TextFooler}
Di~Jin, Zhijing Jin, Joey~Tianyi Zhou, and Peter Szolovits. 2020.
\newblock \href {https://aaai.org/ojs/index.php/AAAI/article/view/6311} {Is
  {BERT} really robust? {A} strong baseline for natural language attack on text
  classification and entailment}.
\newblock In \emph{AAAI}, pages 8018--8025.

\bibitem[{Khashabi et~al.(2020)Khashabi, Min, Khot, Sabharwal, Tafjord, Clark,
  and Hajishirzi}]{UnifiedQA}
Daniel Khashabi, Sewon Min, Tushar Khot, Ashish Sabharwal, Oyvind Tafjord,
  Peter Clark, and Hannaneh Hajishirzi. 2020.
\newblock \href {https://www.aclweb.org/anthology/2020.findings-emnlp.171/}
  {Unifiedqa: Crossing format boundaries with a single {QA} system}.
\newblock In \emph{Findings, {EMNLP}}, pages 1896--1907.

\bibitem[{Kocijan et~al.(2019)Kocijan, Cretu, Camburu, Yordanov, and
  Lukasiewicz}]{SurprisinglyRobustTrick}
Vid Kocijan, Ana{-}Maria Cretu, Oana{-}Maria Camburu, Yordan Yordanov, and
  Thomas Lukasiewicz. 2019.
\newblock \href {https://doi.org/10.18653/v1/p19-1478} {A surprisingly robust
  trick for the winograd schema challenge}.
\newblock In \emph{ACL}, pages 4837--4842.

\bibitem[{Lin et~al.(2019)Lin, Chen, Chen, and Ren}]{KagNet}
Bill~Yuchen Lin, Xinyue Chen, Jamin Chen, and Xiang Ren. 2019.
\newblock \href {https://doi.org/10.18653/v1/D19-1282} {Kagnet: Knowledge-aware
  graph networks for commonsense reasoning}.
\newblock In \emph{EMNLP-IJCNLP}, pages 2829--2839.

\bibitem[{Liu et~al.(2019)Liu, Ott, Goyal, Du, Joshi, Chen, Levy, Lewis,
  Zettlemoyer, and Stoyanov}]{roberta}
Yinhan Liu, Myle Ott, Naman Goyal, Jingfei Du, Mandar Joshi, Danqi Chen, Omer
  Levy, Mike Lewis, Luke Zettlemoyer, and Veselin Stoyanov. 2019.
\newblock \href {http://arxiv.org/abs/1907.11692} {Roberta: {A} robustly
  optimized {BERT} pretraining approach}.
\newblock \emph{CoRR}.

\bibitem[{Ma et~al.(2020)Ma, Ilievski, Francis, Bisk, Nyberg, and
  Oltramari}]{DataConstructionForZeroShotCQA}
Kaixin Ma, Filip Ilievski, Jonathan Francis, Yonatan Bisk, Eric Nyberg, and
  Alessandro Oltramari. 2020.
\newblock \href {https://arxiv.org/pdf/2011.03863.pdf} {Knowledge-driven data
  construction for zero-shot evaluation in commonsense question answering}.
\newblock \emph{CoRR}.

\bibitem[{Moghimifar et~al.(2020)Moghimifar, Qu, Zhuo, Baktashmotlagh, and
  Haffari}]{Cosmo}
Farhad Moghimifar, Lizhen Qu, Yue Zhuo, Mahsa Baktashmotlagh, and Gholamreza
  Haffari. 2020.
\newblock \href {https://doi.org/10.18653/v1/2020.coling-main.467} {Cosmo:
  Conditional seq2seq-based mixture model for zero-shot commonsense question
  answering}.
\newblock In \emph{COLING}, pages 5347--5359.

\bibitem[{Mostafazadeh et~al.(2016)Mostafazadeh, Chambers, He, Parikh, Batra,
  Vanderwende, Kohli, and Allen}]{StoryClozeTest}
Nasrin Mostafazadeh, Nathanael Chambers, Xiaodong He, Devi Parikh, Dhruv Batra,
  Lucy Vanderwende, Pushmeet Kohli, and James Allen. 2016.
\newblock \href {https://www.aclweb.org/anthology/N16-1098} {A corpus and cloze
  evaluation for deeper understanding of commonsense stories}.
\newblock In \emph{NAACL}.

\bibitem[{Pennington et~al.(2014)Pennington, Socher, and Manning}]{GloVe}
Jeffrey Pennington, Richard Socher, and Christopher~D. Manning. 2014.
\newblock \href {https://doi.org/10.3115/v1/d14-1162} {Glove: Global vectors
  for word representation}.
\newblock In \emph{EMNLP}, pages 1532--1543.

\bibitem[{Petroni et~al.(2019)Petroni, Rockt{\"{a}}schel, Riedel, Lewis,
  Bakhtin, Wu, and Miller}]{LMasKB}
Fabio Petroni, Tim Rockt{\"{a}}schel, Sebastian Riedel, Patrick S.~H. Lewis,
  Anton Bakhtin, Yuxiang Wu, and Alexander~H. Miller. 2019.
\newblock \href {https://doi.org/10.18653/v1/D19-1250} {Language models as
  knowledge bases?}
\newblock In \emph{EMNLP-IJCNLP}, pages 2463--2473.

\bibitem[{Radford et~al.(2019)Radford, Wu, Child, Luan, Amodei, and
  Sutskever}]{GPT2}
Alec Radford, Jeffrey Wu, Rewon Child, David Luan, Dario Amodei, and Ilya
  Sutskever. 2019.
\newblock \href
  {https://d4mucfpksywv.cloudfront.net/better-language-models/language-models.pdf}
  {Language models are unsupervised multitask learners}.

\bibitem[{Rajani et~al.(2019)Rajani, McCann, Xiong, and
  Socher}]{ExplainYourself}
Nazneen~Fatema Rajani, Bryan McCann, Caiming Xiong, and Richard Socher. 2019.
\newblock \href {https://doi.org/10.18653/v1/p19-1487} {Explain yourself!
  leveraging language models for commonsense reasoning}.
\newblock In \emph{ACL}, pages 4932--4942.

\bibitem[{Reimers and Gurevych(2019)}]{sentenceBERT}
Nils Reimers and Iryna Gurevych. 2019.
\newblock \href {https://doi.org/10.18653/v1/D19-1410} {Sentence-bert: Sentence
  embeddings using siamese bert-networks}.
\newblock In \emph{EMNLP-IJCNLP}, pages 3980--3990.

\bibitem[{Roemmele et~al.(2011)Roemmele, Bejan, and Gordon}]{COPA}
Melissa Roemmele, Cosmin~Adrian Bejan, and Andrew~S. Gordon. 2011.
\newblock \href {http://www.aaai.org/ocs/index.php/SSS/SSS11/paper/view/2418}
  {Choice of plausible alternatives: An evaluation of commonsense causal
  reasoning}.
\newblock In \emph{AAAI}.

\bibitem[{Sap et~al.(2019{\natexlab{a}})Sap, Bras, Allaway, Bhagavatula,
  Lourie, Rashkin, Roof, Smith, and Choi}]{atomic}
Maarten Sap, Ronan~Le Bras, Emily Allaway, Chandra Bhagavatula, Nicholas
  Lourie, Hannah Rashkin, Brendan Roof, Noah~A. Smith, and Yejin Choi.
  2019{\natexlab{a}}.
\newblock \href {https://doi.org/10.1609/aaai.v33i01.33013027} {{ATOMIC:} an
  atlas of machine commonsense for if-then reasoning}.
\newblock In \emph{AAAI}, pages 3027--3035.

\bibitem[{Sap et~al.(2019{\natexlab{b}})Sap, Rashkin, Chen, Bras, and
  Choi}]{SocialIQA}
Maarten Sap, Hannah Rashkin, Derek Chen, Ronan~Le Bras, and Yejin Choi.
  2019{\natexlab{b}}.
\newblock \href {https://doi.org/10.18653/v1/D19-1454} {Social iqa: Commonsense
  reasoning about social interactions}.
\newblock In \emph{EMNLP}, pages 4462--4472.

\bibitem[{Shwartz et~al.(2020)Shwartz, West, Bras, Bhagavatula, and
  Choi}]{selfTalk}
Vered Shwartz, Peter West, Ronan~Le Bras, Chandra Bhagavatula, and Yejin Choi.
  2020.
\newblock \href {https://arxiv.org/abs/2004.05483v2} {Unsupervised commonsense
  question answering with self-talk}.
\newblock In \emph{EMNLP}.

\bibitem[{Speer et~al.(2017)Speer, Chin, and Havasi}]{ConceptNet}
Robyn Speer, Joshua Chin, and Catherine Havasi. 2017.
\newblock \href {http://aaai.org/ocs/index.php/AAAI/AAAI17/paper/view/14972}
  {Conceptnet 5.5: An open multilingual graph of general knowledge}.
\newblock In \emph{AAAI}, pages 4444--4451.

\bibitem[{Talmor et~al.(2019)Talmor, Herzig, Lourie, and
  Berant}]{CommonsenseQA}
Alon Talmor, Jonathan Herzig, Nicholas Lourie, and Jonathan Berant. 2019.
\newblock \href {https://doi.org/10.18653/v1/n19-1421} {Commonsenseqa: {A}
  question answering challenge targeting commonsense knowledge}.
\newblock In \emph{NAACL-HLT}, pages 4149--4158.

\bibitem[{Tamborrino et~al.(2020)Tamborrino, Pellican{\`{o}}, Pannier, Voitot,
  and Naudin}]{LMisAllYouNeedForCSReasoning}
Alexandre Tamborrino, Nicola Pellican{\`{o}}, Baptiste Pannier, Pascal Voitot,
  and Louise Naudin. 2020.
\newblock \href {https://www.aclweb.org/anthology/2020.acl-main.357/}
  {Pre-training is (almost) all you need: An application to commonsense
  reasoning}.
\newblock In \emph{ACL}, pages 3878--3887.

\bibitem[{Trinh and Le(2018)}]{ASimpleMethod}
Trieu~H. Trinh and Quoc~V. Le. 2018.
\newblock \href {http://arxiv.org/abs/1806.02847} {A simple method for
  commonsense reasoning}.
\newblock \emph{CoRR}.

\bibitem[{Wang et~al.(2020)Wang, Peng, Ilievski, Szekely, and
  Ren}]{PathGenerator}
Peifeng Wang, Nanyun Peng, Filip Ilievski, Pedro~A. Szekely, and Xiang Ren.
  2020.
\newblock \href {https://www.aclweb.org/anthology/2020.findings-emnlp.369/}
  {Connecting the dots: {A} knowledgeable path generator for commonsense
  question answering}.
\newblock In \emph{Findings, {EMNLP}}, pages 4129--4140.

\bibitem[{Wang et~al.(2019)Wang, Zhang, Shen, Liu, Liu, Gao, and Jiang}]{UDSSM}
Shuohang Wang, Sheng Zhang, Yelong Shen, Xiaodong Liu, Jingjing Liu, Jianfeng
  Gao, and Jing Jiang. 2019.
\newblock \href {https://doi.org/10.18653/v1/n19-1094} {Unsupervised deep
  structured semantic models for commonsense reasoning}.
\newblock In \emph{NAACL-HLT}, pages 882--891.

\end{thebibliography}

\newpage
\appendix

\section{Datasets}

The four datasets used in this work are multiple-choice commonsense question answering tasks. 

\href{https://people.ict.usc.edu/~gordon/copa.html}{\textbf{COPA}}\footnote{https://people.ict.usc.edu/~gordon/copa.html}~\citep{COPA} evaluates the ability of causal reasoning about a certain event, which is expressed in a simple sentence.
Each question is accompanied with two candidate choices.

\href{https://www.cs.rochester.edu/nlp/rocstories/}{\textbf{StoryClozeTest (SCT)}}\footnote{https://www.cs.rochester.edu/nlp/rocstories/}~\citep{StoryClozeTest} requires models to select the reasonable story ending, from two alternatives, conditioned on a description about the story context.

\href{https://leaderboard.allenai.org/socialiqa/submissions/get-started}{\textbf{SocialIQA}}\footnote{https://leaderboard.allenai.org/socialiqa/submissions/get-started}~\citep{SocialIQA} evaluates the reasoning ability on social events. In each example, the question describes a social event and asks models to make some inferences based on the event, such as its cause or effect. 

\href{https://leaderboard.allenai.org/cosmosqa/submissions/get-started}{\textbf{CosmosQA}}\footnote{https://leaderboard.allenai.org/cosmosqa/submissions/get-started}~\citep{CosmosQA} is a reading comprehension task. Different from the three datasets above, the examples of CosmosQA have long and complex contexts. The original dataset contains a type of choices ``None of the above'' to test whether models can identify unanswerable questions. This is not the focus of our work, so we removed such choices.

For COPA, we reported the results on its test set.
As the test sets of SCT, SocialIQA and CosmosQA are hidden, for convenience of analysis,
we reported the experiment results on their development sets. 
See Table~\ref{tab:dataset-statistic} for statistic information of each dataset.

\begin{table*}[ht]
\small
\centering
\begin{tabular}{lccccc}\toprule[1pt]
Dataset                     & COPA-dev   & COPA-test  & SCT-dev     & SocialIQA-dev & CosmosQA-dev \\ \hline
Number of Examples          & 500        & 500        & 1571        & 1954          & 2726         \\
Number of Choices           & 2          & 2          & 2           & 3             & 3/4          \\
Question Length (mean, std) & (7.3, 1.8) & (7.1, 1.7) & (35.3, 6.5) & (15.3, 4.4)   & (83.0, 24.5) \\
Choice Length (mean, std)   & (5.1, 1.6) & (5.0, 1.5) & (7.4, 2.5)  & (3.7, 2.3)    & (10.0, 4.3) 
\\\bottomrule[1pt]
\end{tabular}
\caption{Statistic information of each dataset. Due to the removal of the choice ``None of the above'', each instance of CosmosQA may have 3 or 4 answer choices.}
\label{tab:dataset-statistic}
\end{table*}

\section{Templates for Rewriting Questions}

We use the same templates for our method and all the baselines. Note that the templates for rewriting questions is not the focus of this paper, and we inherit the templates from previous work if available.

\citet{LMisAllYouNeedForCSReasoning} provides templates for \textbf{COPA} (Table~\ref{tab:convert-question-copa}) and \citet{selfTalk} provides templates for \textbf{SocialIQA} (Table~\ref{tab:convert-question-socialiqa}). Since the instances in \textbf{SCT} have no questions, \textbf{SCT} does not need templates. There is no related work discussing templates for \textbf{CosmosQA}, so we design some templates by ourselves (Table~\ref{tab:convert-question-cosmosqa}).
\textbf{Source code for rewriting questions and \methodname will be made publicly available.}

\begin{table*}[t]
\centering
\begin{tabular}{p{0.41\linewidth}p{0.41\linewidth}}\toprule[1pt]
\textbf{Original Question}                                                                                          & \textbf{Rewrite}                                                                               \\ \hline
What was the cause of this?                                                                                         & because                                                                                        \\
What happened as a result?                                                                                          & so                                                                                             \\\hline\hline
\textbf{Original Example}                                                                                           & \textbf{Rewrite}                                                                               \\ \hline
I saw my breath when I exhaled. {\color{red} What was the cause of this?} The weather was chilly. & I saw my breath when I exhaled {\color{red} because} the weather was chilly.
\\\bottomrule[1pt]
\end{tabular}
\caption{Templates and a rewritten example of COPA. The templates are inherited from \citet{LMisAllYouNeedForCSReasoning}.}
\label{tab:convert-question-copa}
\end{table*}

\begin{table*}[t]
\small
\centering
\begin{tabular}{p{0.3\linewidth}p{0.3\linewidth}p{0.3\linewidth}}\toprule[1pt]
\textbf{Original Question}                                                                                                                   & \textbf{Rewrite 1}                                                                                                                        & \textbf{Rewrite 2}                                                                                                                          \\ \hline
What will {[}SUBJ{]} want to do next?                                                                                                        & As a result, {[}SUBJ{]} wanted to                                                                                                         & \textless{}xwant\textgreater{}                                                                                                              \\
How would {[}SUBJ{]} feel as a result?                                                                                                       & As a result, {[}SUBJ{]} felt                                                                                                              & \textless{}xeffect\textgreater{}                                                                                                            \\
What will {[}SUBJ{]} do next?                                                                                                                & {[}SUBJ{]} then                                                                                                                           & \textless{}xreact\textgreater{}                                                                                                             \\
How would you describe {[}SUBJ{]}?                                                                                                           & {[}SUBJ{]} is seen as                                                                                                                     & \textless{}xattr\textgreater{}                                                                                                              \\
Why did {[}SUBJ{]} do that?                                                                                                                  & Before, {[}SUBJ{]} wanted                                                                                                                 & \textless{}xintent\textgreater{}                                                                                                            \\
What does {[}SUBJ{]} need to do before?                                                                                                      & Before, {[}SUBJ{]} needed to                                                                                                              & \textless{}xneed\textgreater{}                                                                                                              \\
\hline\hline
\textbf{Original Example}                                                                                                                    & \textbf{Rewrite 1}                                                                                                                        & \textbf{Rewrite 2}                                                                                                                          \\ \hline
Sydney went trick or treating and the others joined him happily. {\color{red}What will Others want to do next?} get candy & Sydney went trick or treating and the others joined him happily. {\color{red}As a result, Others wanted to} get candy. & Sydney went trick or treating and the others joined him happily. {\color{red} \textless{}xwant\textgreater{}} get candy.
\\\bottomrule[1pt]
\end{tabular}
\caption{Some templates and a rewritten example of SocialIQA. {[}SUBJ{]} refers to a subject. There are two groups of templates, Rewrite1 for GPT-2 and Rewrite2 for COMET~\citep{COMET}. The relations in Rewrite2 are defined in \citet{atomic} and used for training COMET. These templates are inherited from \citet{selfTalk}. More details can be found in \citet{selfTalk} and \href{https://github.com/vered1986/self_talk}{https://github.com/vered1986/self\_talk}.}
\label{tab:convert-question-socialiqa}
\end{table*}

\begin{table*}[t]
\centering
\begin{tabular}{p{0.45\linewidth}p{0.45\linewidth}}\toprule[1pt]
\textbf{Original Question}                                                                                                                                                                                                                                & \textbf{Rewrite}                                                                                                                                                                                                                                             \\ \hline
Why {[}SENTENCE{]} {[}CLAUSE{]} ?                                                                                                                                                                                                                         & {[}CLAUSE{]} {[}SENTENCE{]} because                                                                                                                                                                                                                          \\
What {[}NOUN{]} {[}SENTENCE{]} {[}CLAUSE{]} ?                                                                                                                                                                                                             & {[}CLAUSE{]} the {[}NOUN{]} {[}SENTENCE{]} is that                                                                                                                                                                                                           \\
What {[}SENTENCE{]} {[}CLAUSE{]} ?                                                                                                                                                                                                                        & {[}CLAUSE{]} it {[}SENTENCE{]} that                                                                                                                                                                                                                          \\\hline\hline
\textbf{Original Example}                                                                                                                                                                                                                                 & \textbf{Rewrite}                                                                                                                                                                                                                                             \\ \hline
... He was conscious but seemed dazed and probably intoxicated . Nearby there was a young man dialing his cell phone . {\color{red}What may happen after the young man makes his call ?} An ambulance would likely come to the scene . & ... He was conscious but seemed dazed and probably intoxicated . Nearby there was a young man dialing his cell phone . {\color{red}After the young man makes his call , it may happen that} an ambulance would likely come to the scene .
\\\bottomrule[1pt]
\end{tabular}
\caption{Templates and a rewritten example of CosmosQA. {[}NOUN{]}, {[}SENTENCE{]} and {[}CLAUSE{]} refer to a noun, a sentence fragment and an adverbial clause, respectively.}
\label{tab:convert-question-cosmosqa}
\end{table*}

\section{Selection of Pre-trained Models}

For each method, we tried to adopt different pre-trained models and find the pre-trained models that maximized the accuracy on the development set of each dataset. 
Table~\ref{tab:selection-of-pretrained-models} shows the set of candidate pre-trained models for each method, with the selected models in bold. Because of the nature of Pro-Q, it can only use bidirectional language models, so we only evaluated Pro-Q with RoBERTa-large and SentenceRoBERTa-large.

As shown in Table~\ref{tab:selection-of-pretrained-models}, for each method except CGA, the best selection of pre-trained models is consistent on all the datasets. CGA achieved its best performance with COMET on SocialIQA and with GPT2-xlarge on the other datasets.

\begin{table*}[t]
\centering
\begin{tabular}{ll}\toprule[1pt]
Method    & Set of Candidate Pre-trained Models                                                                                      \\ \hline
Pro-A     & LM as QA model: (\textbf{GPT2-xlarge}, COMET, RoBERTa-large, SentenceRoBERTa-large)                                                                                     \\ \hline
Pro-Q     & LM as QA model: (\textbf{RoBERTa-large}, SentenceRoBERTa-large)                                                                                         \\ \hline
MI-QA     & LM as QA model: (\textbf{GPT2-xlarge}, COMET, RoBERTa-large, SentenceRoBERTa-large)                                                                                     \\ \hline
Self-talk & \begin{tabular}[c]{@{}l@{}}LM as generator: (GPT2-xlarge, \textbf{COMET})\\ LM as QA model: (\textbf{GPT2-xlarge}, COMET, RoBERTa-large, SentenceRoBERTa-large)\end{tabular}     \\ \hline
CGA       & LM as QA model and generator: (\textbf{GPT2-xlarge}, \textbf{COMET})                                                                       \\ \hline
SEQA      & \begin{tabular}[c]{@{}l@{}}LM as generator: (\textbf{GPT2-xlarge}, COMET)\\ Feature Extractor: \textbf{SentenceRoBERTa-large}\end{tabular}
\\\bottomrule[1pt]
\end{tabular}
\caption{The set of candidate pre-trained models. The selected pre-trained models for each method are marked in bold. Note that CGA achieved its best performance with COMET on SocialIQA and with GPT2-xlarge on the other datasets.}
\label{tab:selection-of-pretrained-models}
\end{table*}

\section{Hyperparameter Search}

\begin{table}[!t]
\centering
\begin{tabular}{lcc}\toprule[1pt]
$T$  & Dev & Test \\ \hline
10   &  70.0   &   75.6   \\
1    &  70.4   &   76.4   \\
0.2  &  71.8   &   77.0   \\
0.1  &  \textbf{75.4}   &   79.4   \\
0.05 &  74.4   &   80.2  
\\\bottomrule[1pt]
\end{tabular}
\caption{Hyperparameter Search of \methodname. The temperature is selected according to the accuracy on the development set of COPA.}
\label{tab:hyperparameter-search}
\end{table}

For \methodname, we only tuned the temperature $T$.
To be more specific, we selected $T$ from five candidate values according to the accuracy on the development set of COPA. Table~\ref{tab:hyperparameter-search} shows that \methodname with $T=0.1$ achieved the best performance on the development set of COPA.
And then we evaluated \methodname with $T=0.1$ on the test set of COPA as well as the development sets of SCT, SocialIQA and CosmosQA.

\section{Analysis on Sample Size}

\begin{figure}[ht]
    \centering
    \includegraphics[width=\linewidth]{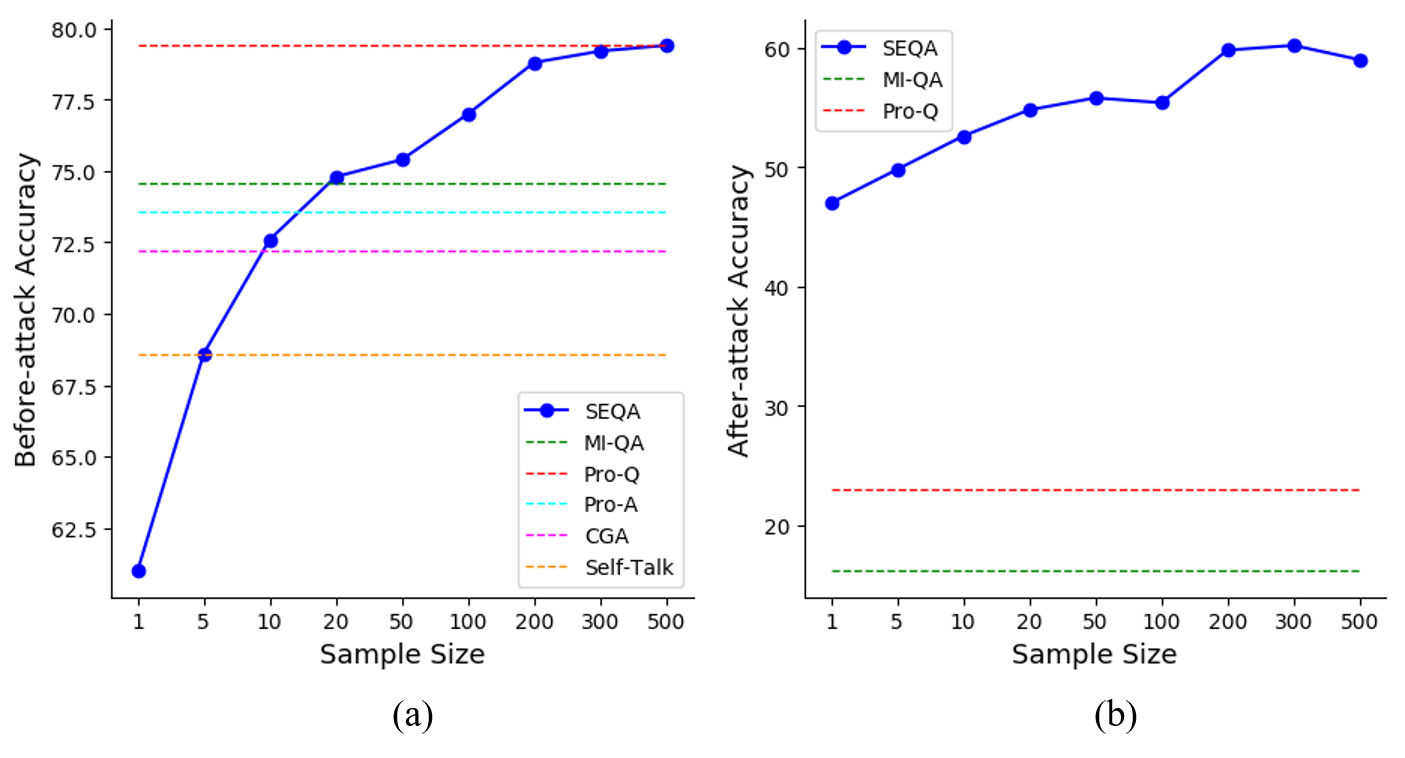}
    \caption{The before-attack (a) and after-attack accuracy (b) of methods with different sample sizes on COPA. The after-attack accuracy of Pro-A, CGA and Self-Talk is below 10.0\%, and thus omitted in (b).}
    \label{fig:ablation-sample-number-COPA}
\end{figure}

\begin{figure}[p]
    \centering
    \includegraphics[width=\linewidth]{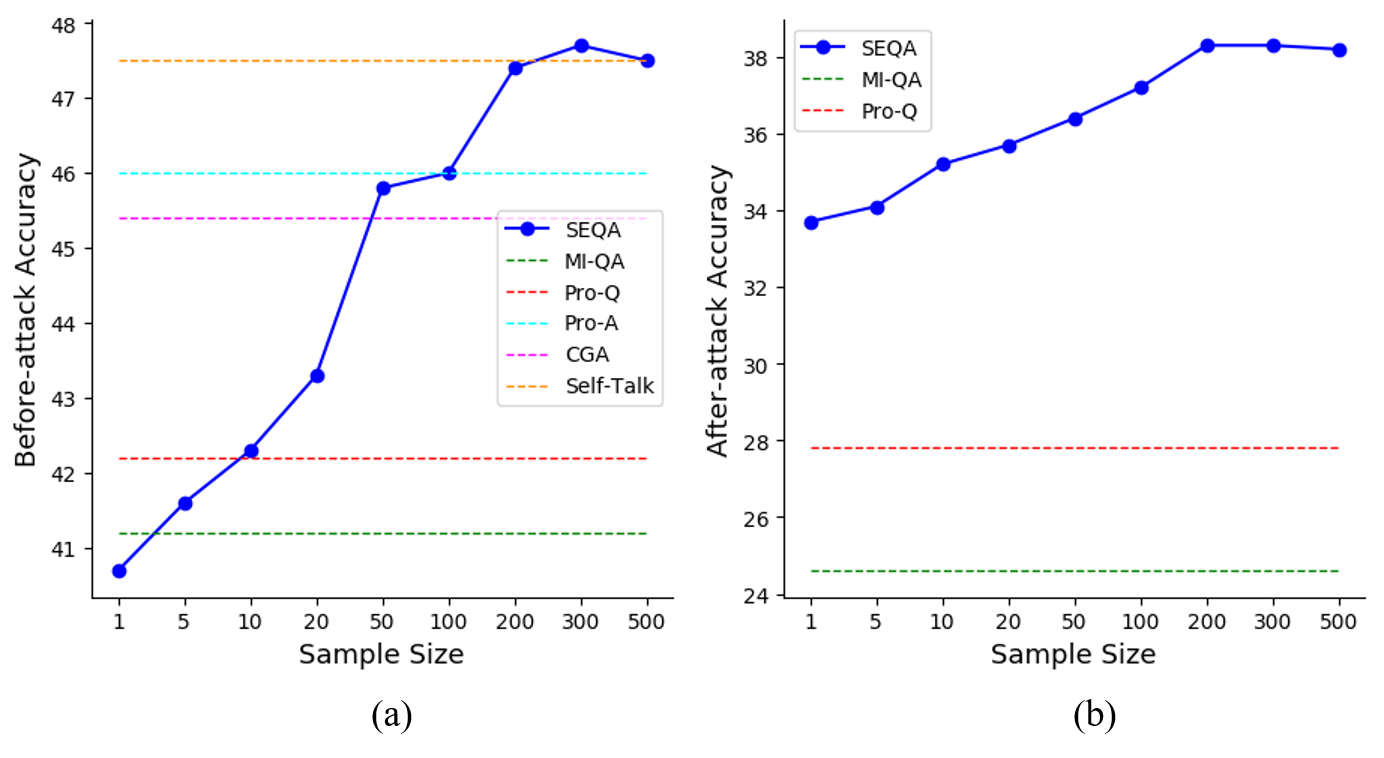}
    \caption{The before-attack (a) and after-attack accuracy (b) of methods with different sample sizes on SocialIQA. The after-attack accuracy of Pro-A, CGA and Self-Talk is below 20.0\%, and thus omitted in (b).}
    \label{fig:ablation-sample-number-SocialIQA}
\end{figure}

\begin{figure}[p]
    \centering
    \includegraphics[width=\linewidth]{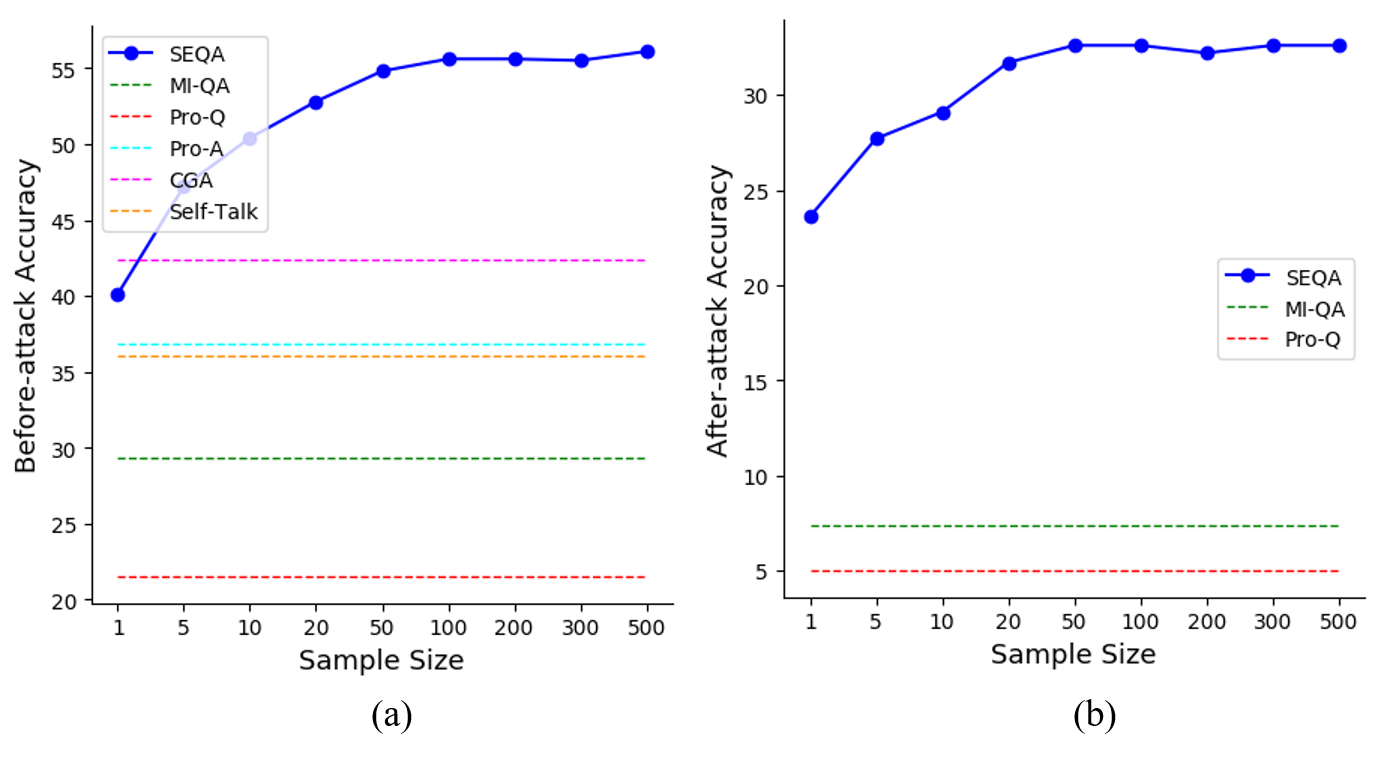}
    \caption{The before-attack (a) and after-attack accuracy (b) of methods with different sample sizes on CosmosQA. The after-attack accuracy of Pro-A, CGA and Self-Talk is below 2.0\%, and thus omitted in (b).}
    \label{fig:ablation-sample-number-CosmosQA}
\end{figure}

Figure~\ref{fig:ablation-sample-number-COPA},\ref{fig:ablation-sample-number-SocialIQA},\ref{fig:ablation-sample-number-CosmosQA} shows the effect of the sample size $K$ on \methodname.
For comparison, these figures also include the results of baselines in the settings of before- and after-attack, respectively.
On the overall trend, the performance of \methodname improved as the sample size increased.
Another observation is that a smaller sample size can already make \methodname outperform most baseline methods.

\pagebreak

\end{document}